\pdfoutput=1

\documentclass[11pt]{article}
\PassOptionsToPackage{hyperfootnotes=false}{hyperref}

\usepackage{acl}

\usepackage{times}
\usepackage{latexsym}

\usepackage[T1]{fontenc}

\usepackage[utf8]{inputenc}

\usepackage{microtype}

\usepackage{inconsolata}

\usepackage{hyperref}
\usepackage{url}
\usepackage{booktabs}
\usepackage{amsfonts}
\usepackage{nicefrac}
\usepackage{xcolor}
\usepackage{amsmath}
\usepackage{amssymb}
\usepackage[section]{placeins}
\usepackage{graphicx}
\usepackage{natbib}
\usepackage{tabularx}
\usepackage{xspace}
\usepackage{colortbl}
\usepackage{soul}
\usepackage{stfloats}
\usepackage{algorithm,algorithmicx}
\usepackage[noend]{algpseudocode}
\usepackage{algcompatible}
\usepackage{pifont}
\usepackage{makecell}
\usepackage{siunitx}
\usepackage{subcaption}
\graphicspath{{./figures/}}
\usepackage{multirow}

\definecolor{lightgrey}{HTML}{dcdbdb}
\sethlcolor{lightgrey}
\definecolor{lightblue}{HTML}{E8F0FE}
\definecolor{lightblue}{HTML}{E8F0FE}
\definecolor{gray}{HTML}{9aa0a6}
\definecolor{lightpink}{HTML}{F48FB1}
\definecolor{lightred}{HTML}{FFCBC9}
\definecolor{lightcyan}{HTML}{80DEEA}

\usepackage[skins]{tcolorbox} 
\tcbuselibrary{breakable} 
\newtcolorbox[auto counter, number within=section, list type=subsubsection, list inside=toc]{sectionbox}[2][]{
colback=white!98!gray, colframe=black, 
colbacktitle=white!90!gray, coltitle=black, 
fonttitle=\bfseries,
title={#2}, 
list entry={Comment \thetcbcounter\quad}
}

\newcommand{\ourmodel}{Reflective Recovery\xspace}
 

\title{%
    \ourmodel: A Self-Supervised Method for Reasoning by Learning from Mistakes
}

\author{Qirui Chen$^*$ \\
  Zhejiang University \\
  The University of Hong Kong \\
  \texttt{3220103318@zju.edu.cn} \\\And
  Renjie Pi$^*$ \\
  The Hong Kong University of \\
  Science and Technology \\
  \texttt{rpi@connect.ust.hk} \\\AND
  Jiahui Gao$^\dagger$ \\
  The University of Hong Kong \\
  \texttt{ggaojiahui@gmail.com} \\\And
  Lingpeng Kong \\
  The University of Hong Kong \\
  \texttt{lpk@cs.hku.hk} \\}
\begin{document}

\maketitle
{\renewcommand{\thefootnote}{}\footnotetext{$^*$Equal contribution. \quad $^\dagger$Corresponding author.}}

\begin{abstract}
Data-driven fine-tuning is widely adopted to enhance reasoning in Large Language Models (LLMs) due to its simplicity and efficiency. However, mainstream imitation learning methods that rely exclusively on perfect reasoning trajectories suffer from a \textit{Scaling Collapse}: when the problem set is limited, increasing positive examples fails to yield continuous improvement. However, during inference, an LLM can not guarantee that every intermediate step is correct and is therefore prone to errors. Once such errors arise, the LLM often struggles to recover and may be further misled by the accumulation of previous mistakes. To address this, we propose \textbf{Reflective Recovery}, a simple yet effective self-supervised approach that transforms failed reasoning attempts into recovery training data. Specifically, we extract initial segments of failed trajectories, concatenate them with prompts, and use them to guide the LLM toward valid solutions. 
Because these segments from failed trajectories are likely to contain errors, this process teaches models to recognize and correct mistakes during reasoning, enabling recovery from erroneous states without relying on external critics or reward models.
Evaluated on extensive benchmarks, Reflective Recovery significantly improves performance. On DeepSeek-R1-Distill-Qwen-7B, it boosts accuracy from 30.0\% to 37.5\% on AIME 2025 and from 37.6\% to 47.8\% on Minerva.
More importantly, analyses demonstrate that it breaks the scaling collapse barrier and enables models to develop emergent self-correction behaviors, representing a paradigm shift from outcome-oriented memorization to process-oriented reflective reasoning.
\end{abstract}
\section{Introduction}
Large Language Models (LLMs) have achieved remarkable progress across diverse natural language tasks, from question answering to code generation~\citep{brown2020languagemodelsfewshotlearners, grattafiori2024llama3herdmodels}. Among these capabilities, complex reasoning stands out as a critical frontier for tackling multi-step mathematical problems and scientific challenges~\citep{openai2024openaio1card, deepseekai2025, Anthropic@claude}. Consequently, enhancing the reasoning capacity of LLMs has gained attention from both academia and industry.

Among various approaches to improving reasoning ability (e.g., supervised fine-tuning, reinforcement learning), data-driven fine-tuning has emerged as the most prevalent solution due to its computational efficiency and simplicity in training.
Most recent progress~\citep{chen2025longpolongcontextselfevolution,liu2025rss,gulcehre2023reinforced, yang2024supercorrect} follows an imitation learning paradigm, with Rejection Sampling Fine-Tuning (RFT) being the most representative. RFT collects high-quality, correct reasoning trajectories from large-scale sampling as positive examples for training~\citep{yuan2023scalingrelationshiplearningmathematical}. 
During the initial alignment phase, this imitation approach can achieve substantial performance improvements with a moderate amount of data.

However, our experiments reveal a fundamental efficiency bottleneck in this positive-only imitation learning approach. When the problem set is limited, merely increasing the sampling volume to accumulate more positive examples does not continuously translate into improved model capabilities. For instance, on the AIME 2025 benchmark, the RFT baseline's performance actually degrades from 35.3\% to 30.0\% as the training data volume scales from 8k to 10k. Our experiments show that as the scale of positive training data increases, performance gains rapidly diminish and eventually stagnate—a phenomenon we term \textit{scaling collapse} (Figure~\ref{fig:all_pass1_datasets_alt}). Despite being exposed to more correct paths, models still struggle with complex, multi-step problems. This exposes a fundamental limitation of imitation learning: it guides models to memorize correct reasoning trajectories rather than develop robust reasoning strategies that generalize beyond memorized patterns.

The limitation of this approach becomes particularly significant when models deviate from the correct trajectory: since the training data never includes error states, once a model makes an incorrect reasoning step, it lacks any mechanism for self-correction. Effective reasoning should be a dynamic process that requires not only generating correct logic, but also identifying and correcting errors when they arise.
To address this, recent work has explored mechanisms that encourage recovery or self-correction.
However, approaches relying on expensive annotations or additional models for external feedback~\citep{kumar2024training, wang2025critiquefinetuninglearningcritique} suffer from limited scalability and autonomy, while methods prompting models for intrinsic reflection after complete failure~\citep{shinn2023reflexion} lack the capability to perform mid-process correction during reasoning.


In this work, we explore a simpler alternative: rather than introducing new critics or supervision signals, we investigate whether models can learn to recover directly from their own failures. To this end, we introduce \textbf{Reflective Recovery}, a self-supervised framework that constructs recovery training data from failed reasoning attempts.
Specifically, given a set of negative reasoning trajectories ($\mathcal{D}_{\text{neg}}$) that lead to incorrect answers, we extract their error prefixes—partial reasoning chains. We then pair each error prefix with a correct continuation that leads to the right solution. This process yields a reflective dataset ($\mathcal{D}_{\text{rec}}$) where each training instance consists of an error state and its corresponding recovery path. By training on such data, models learn to identify and correct reasoning errors, moving from incorrect intermediate states to correct solutions.

Despite its simplicity, Reflective Recovery does not require external reward models or reinforcement learning, making it highly practical and scalable. Experiments on challenging benchmarks such as AIME and Minerva show that our method outperforms outcome-based RFT baselines. More importantly, it exhibits stable and robust scaling behavior, avoiding the performance saturation observed in standard imitation learning (Figure~\ref{fig:all_pass1_datasets_alt}). Specifically, while the RFT baseline suffers a performance drop on AIME 2025 from 35.3\% to 30.0\% as training data scales from 8k to 10k, our method maintains a steady improvement on benchmarks like Minerva, demonstrating generalization ability.

The main contributions of this work are:
\begin{itemize}
\item We identify and characterize the \textit{scaling collapse} phenomenon in imitation learning, demonstrating that its root cause lies in the absence of error recovery capability.
\item We introduce Reflective Recovery, a simple yet effective self-supervised framework that transforms negative reasoning trajectories into high-value training data for learning recovery behavior.
\item Extensive experiments demonstrate that our method significantly outperforms RFT across multiple challenging benchmarks. Critically, unlike RFT which suffers from performance saturation, Reflective Recovery maintains robust linear scaling behavior with increasing data scale (Figure~\ref{fig:all_pass1_datasets_alt}).
\item Through attribution analysis, we demonstrate that our method elicits dynamic reflective reasoning patterns in models—actively detecting errors and correcting reasoning paths during the inference process (Section~\ref{sec:reflective_analysis}).
\end{itemize}

\section{Related Work}
\paragraph{External Feedback.}

A direction to enhance LLM reasoning is guiding the generation process using external supervision signals. This typically involves the use of additional components or signals to align model outputs with correctness criteria, such as reward models or verifiers~\citep{kumar2024training, feyza2024rl4f, zhang2024generative, wang2025critiquefinetuninglearningcritique, jiang2024intrinsicselfcorrectionenhancementmonte, fajardo2025reflect, zhu2025surprisingeffectivenessnegativereinforcement, an2024learningmistakesmakesllm}. Such feedback can be applied to evaluate reasoning steps or answers. By leveraging these external critics, models can reject incorrect paths or be fine-tuned by reinforcement learning to maximize rewards.

Recent advances has moved towards providing feedback at each step of the reasoning process rather than only on the final answer. Process Reward Models (PRMs) offer supervision on intermediate steps to guide the model toward correct reasoning~\citep{pang2024stepkto, fajardo2025reflect, luo2025reflectevo}. This step-level feedback also supports the automatic generation of reasoning data, where methods like SRA-MCTS and GFlowNet use simple verifiers to explore different reasoning paths and collect higher-quality training examples~\citep{wang2025sra, liu2025rss}. However, these approaches still rely on carefully designed critics or large amounts of human preference data. The need for strong external supervision and costly training procedures limits their scalability~\citep{valmeekam2024small,zhang2025processbasedselfrewardinglanguagemodels}, motivating the development for methods that can operate with minimal external guidance.

\paragraph{Self-Correction and Learning from Errors.}

Parallel to external supervision, intrinsic self-correction focuses on the model's ability to refine its own outputs. Early inference-time approaches such as Reflexion~\citep{shinn2023reflexion} prompt models to verbally critique their answers. However, prior studies show that LLMs often have difficulty recognizing their own mistakes, particularly in multi-step or complex reasoning tasks~\citep{huang2025selfincorrect}, and self-correction can even reduce performance on more challenging benchmarks~\citep{zeng2025correctbench, lee2024volcano}.

To overcome these limitations, recent work has shifted toward training-based correction using negative signals. Methods like SuperCorrect and RISE fine-tune models on iterative refinement traces~\citep{yang2024supercorrect, qu2024recursive, biggs2025learning}. Other approaches train models to avoid incorrect behaviors or learn from corrected reasoning paths, which has been shown to improve robustness~\citep{pi2024strengtheningmultimodallargelanguage, chen2025longpolongcontextselfevolution,zhu2025surprisingeffectivenessnegativereinforcement, zhang2025gpo, li2025enhancingreasoningprocesssupervision, bui2025neintellingdontwant, liu2025noisyrolloutreinforcingvisualreasoning, xiong2025selfrewardingcorrectionmathematicalreasoning, zelikman2022starbootstrappingreasoningreasoning, shao2025deepseekmathv2selfverifiablemathematicalreasoning}. Despite these advances, most approaches remain post-hoc or outcome-focused. It means they intervene only after an error has fully occurred or simply discourage incorrect outputs. As a result, they do not enable models to actively detect and recover from mistakes during the reasoning process itself.

\section{Methodology}
Our approach \textbf{Reflective Recovery} is based on the idea that robust reasoning requires not only learning from correct solutions, but also learning how to recover from intermediate errors. Instead of discarding failed reasoning paths, we argue that these failures contain rich information and view them as valuable training signals. Building on this insight, we introduce a self-supervised Sample-Resample paradigm that leverages the model’s own failure trajectories as a learning resource to explicitly teach the model how to correct itself. 

\begin{figure*}[t!]
    \centering
    \includegraphics[width=0.8\textwidth]{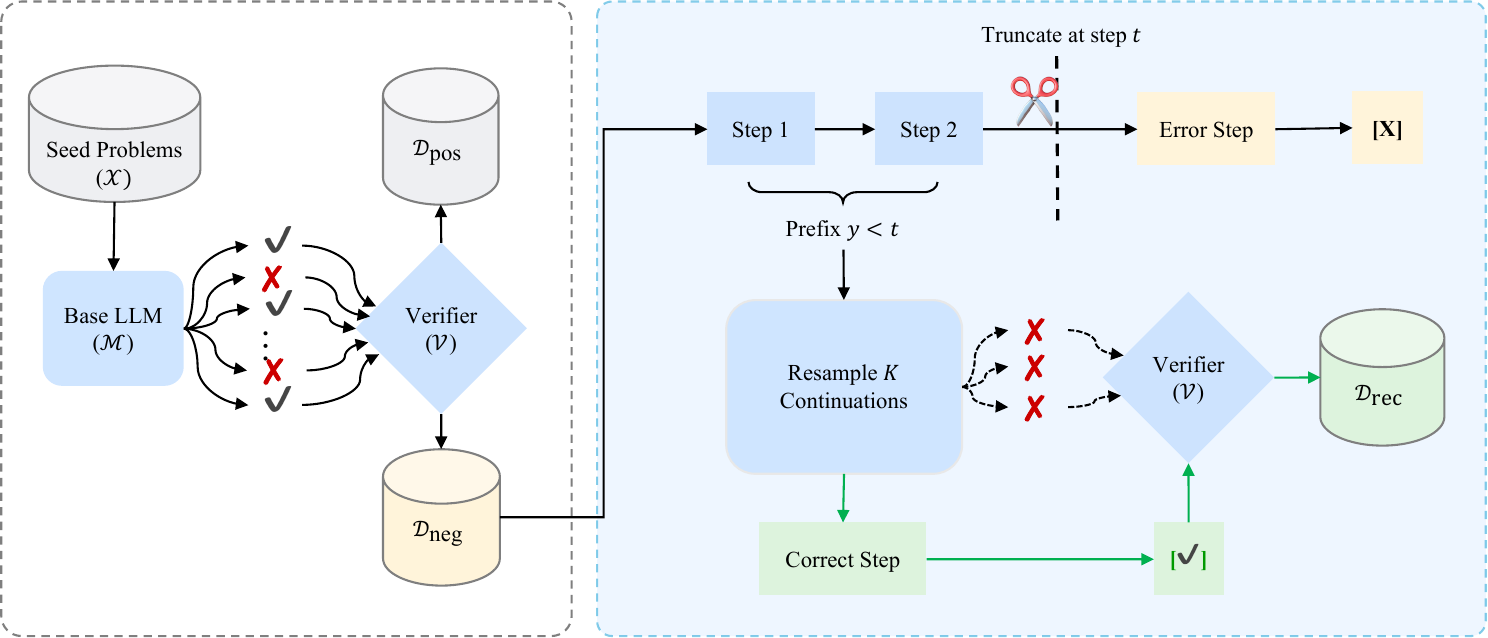}
    \vspace{-2mm}
    \caption{\textbf{The Reflective Recovery pipeline.} We proceed in two phases: (1) \textit{Failure Trajectory Collection}. We apply diverse sampling to extract failure trajectories ($\mathcal{D}_{\text{neg}}$); (2) \textit{Recovery Curation}. We truncate these errors at intermediate step $t$ and resampling to discover valid recovery paths. The resulting dataset ($\mathcal{D}_{\text{rec}}$) contains self-corrected trajectories that enable the model to learn recovery from its own mistake.}
    \label{fig:pipeline}
\end{figure*}

As illustrated in Figure~\ref{fig:pipeline}, our method consists of two phases: (1) extract diverse reasoning trajectories to identify failure trajectories, preserving them as potential recovery opportunities, and (2) transforming failed trajectories into successful, self-corrected trajectories to build a recovery-focused training dataset $\mathcal{D}_{\text{rec}}$.

\subsection{Failure Trajectory Collection}

To enable recovery learning, we first collect a dataset of the model’s mistakes. Let $\mathcal{M}$ denote the base language model and $\mathcal{X}$ be a set of seed problems. In the exploratory phase, for each problem $x \in \mathcal{X}$, we generate $N$ reasoning trajectories $\{y_1, y_2, \dots, y_N\}$ using sampling decoding. By using a high temperature and leveraging the model's uncertainty, we collect failed trajectories that contain diverse errors.

Next, we employ a binary outcome verifier $\mathcal{V}(x, y) \in \{0, 1\}$ to assess the correctness of each trajectory. Based on the verification result, we partition trajectories into a set of correct solutions $\mathcal{D}_{\text{pos}} = \{(x, y) \mid \mathcal{V}(x, y) = 1\}$, and a set of failed trajectories $\mathcal{D}_{\text{neg}} = \{(x, y) \mid \mathcal{V}(x, y) = 0\}$. 

Unlike standard outcome-based methods, which typically discard $\mathcal{D}_{\text{neg}}$, we keep these failure trajectories because they reveal the model's specific error patterns. By studying how the model fails, we can explicitly teach it how to recover from mistakes.

\subsection{Recovery Curation}
The core idea of Reflective Recovery is the converting failed trajectories $\mathcal{D}_{\text{neg}}$ into a dataset of successful recovery trajectories $\mathcal{D}_{\text{rec}}$. 
We achieve this through 
\textbf{Truncate-Resample-Select} procedure.

\paragraph{Truncation.}
For each failed trajectory $y \in \mathcal{D}_{\text{neg}}$, where $y = (y_1, \dots, y_L)$, we cut it off at a intermediate step $t < L$. This yields a partial prefix $y_{<t} = (y_1, \dots, y_{t-1})$. This prefix represents a state where the model might have made mistakes. By stopping here, we create a situation where the model needs to decide how to continue. The selection of $t$ is a hyperparameter controlling the difficulty of recovery (analyzed in Section~\ref{sec:truncation_ablation}).

\paragraph{Resampling for Recovery.}
Given the prefix $(x, y_{<t})$, we sample $K$ alternative continuations $\{\tilde{y}_1, \dots, \tilde{y}_K\}$ from the model using temperature-based sampling. This resampling process explores diverse continuation paths from the same intermediate state, yielding trajectories that may either reach the correct solution or persist in error. The diversity of these samples enables the model to discover multiple recovery strategies from erroneous states.

\paragraph{Verification and Selection.}
Finally, we verify each completion $y_{\text{comp}} = y_{<t} \oplus \tilde{y}_k$ (where $\oplus$ denotes sequence concatenation) using the verifier $\mathcal{V}$. Only trajectories that successfully reach correct solutions are retained to form the recovery dataset.

The resulting recovery dataset $\mathcal{D}_{\mathrm{rec}}$ is formally defined as:
\begin{equation*}
\mathcal{D}_{\mathrm{rec}} = \{\, (x, y_{<t} \oplus \tilde{y}_k) \;\mid\; \mathcal{V}(x, y_{<t} \oplus \tilde{y}_k) = 1 \,\} 
\end{equation*}
Unlike regular datasets that contain only complete correct solutions, $\mathcal{D}_{\mathrm{rec}}$ consists of trajectories that transition from intermediate erroneous states to correct final answers, explicitly capturing the recovery process.

\subsection{Training Objective}
We train the model on the recovery dataset $\mathcal{D}_{\mathrm{rec}}$ to explicitly teach it how to recover from intermediate errors. We employ a specialized supervised fine-tuning objective that focuses exclusively on the recovery completions. For each recovery trajectory $(x, y_{\text{rec}}) \in \mathcal{D}_{\mathrm{rec}}$, where $y_{\text{rec}} = y_{<t} \oplus \tilde{y}_k$, we treat the problem description $x$ and the erroneous prefix $y_{<t}$ as fixed context.
Gradients are only computed for resampled continuation $\tilde{y}$. The training objective minimizes the negative log-likelihood over the tokens in $\tilde{y}$. Formally, for a completion $\tilde{y}$ consisting of tokens $(\tilde{y}_1, \dots, \tilde{y}_m)$, the loss is defined as:
\begin{equation*}
\mathcal{L}(\theta) = - \sum_{j=1}^{m} \log P_{\theta}(\tilde{y}_j \mid x, y_{<t}, \tilde{y}_{<j})
\end{equation*}

By masking the loss for the prefix $y_{<t}$, the model avoids reinforcing its prior mistakes. Instead, it learns to generate the correct continuation $\tilde{y}$ conditional on erroneous trajectory. This training objective enables the model to not only learn correct solutions but also develop the ability to recognize, adapt, and recover from its own mistakes.

\section{Experiments}
\subsection{Setup}

\paragraph{Models and Data.}
Our experiments utilize the DeepSeek-R1-Distill-Qwen family~\citep{deepseekai2025}, specifically the 7B and 14B variants, to evaluate performance scalability. For the seed prompts, we employ 3k high-quality mathematical problems sourced from the SFT Stage 2 dataset of Light-R1~\citep{wen2025lightr1curriculumsftdpo}. These problems are collected from public datasets and have been carefully filtered to remove overlap with evaluation benchmarks such as AIME and GPQA Diamond. This dataset focuses on high-difficulty questions, and each one  needs careful thinking.

\paragraph{Baselines.}
We benchmark our approach against Rejection Sampling Fine-Tuning (RFT)~\citep{yuan2023scalingrelationshiplearningmathematical}. RFT represents the standard outcome-based supervision paradigm, where the model is fine-tuned exclusively on $\mathcal{D}_{\text{pos}}$, which are trajectories yielding the correct final answer. In contrast, our method employs process-based recovery supervision, training on $\mathcal{D}_{\text{rec}}$ derived from corrected failures. To ensure a fair comparison, we align the training data volume for both methods in all experiments, ensuring $|\mathcal{D}_{\text{pos}}| = |\mathcal{D}_{\text{rec}}|$.

\begin{table*}[t]
\centering
\caption{Main experimental results (\texttt{pass@1} accuracy) on DeepSeek-R1-Distill-Qwen 7B and 14B models. All RFT and Ours models are trained on same number of trajectories. The best result in each column is highlighted in \textbf{bold}.}
\label{tab:main-results}
\setlength{\tabcolsep}{1.5pt}
\renewcommand{\arraystretch}{1.25}
\resizebox{\textwidth}{!}{%
\begin{tabular}{l c c c c c c c}
    \toprule
    \textbf{Method} & \textbf{AIME 2024} & \textbf{AIME 2025} & \textbf{LiveCodeBench} & \textbf{GPQA} & \textbf{LiveCodeBench V4} & \textbf{OlympiadBench} & \textbf{Minerva} \\
    \midrule
    \cmidrule(lr){1-8}
    DeepSeek-R1-Distill-Qwen-7B       & 53.0            & 30.0            & 34.4          & 49.1          & 38.4             & 56.4            & 37.6             \\
    + RFT                             & 54.3            & 34.7            & 35.0          & 49.7          & 40.5             & 57.1            & 46.7             \\
    + Reflective Recovery             & \textbf{56.7}   & \textbf{37.5}   & \textbf{36.5} & \textbf{50.3} & \textbf{41.2}    & \textbf{58.1}   & \textbf{47.8}    \\
    \midrule
    \cmidrule(lr){1-8}
    DeepSeek-R1-Distill-Qwen-14B      & 66.6            & 40.0            & 48.3          & 57.5          & 51.5             & 59.0            & 43.8             \\
    + RFT                             & 69.4            & 43.3            & 53.1          & 59.4          & 53.5             & 59.1            & 47.4             \\
    + Reflective Recovery             & \textbf{69.7}   & \textbf{45.3}   & \textbf{55.3} & \textbf{60.3} & \textbf{54.2}    & \textbf{59.4}   & \textbf{50.7}    \\
    \bottomrule
\end{tabular}
}
\end{table*}

\begin{table*}[t!]
    \centering
    \caption{Ablation study on the data truncation position for the DeepSeek-R1-Distill-Qwen 7B model. $L$ denotes the total length of the failed trajectory. The best results are \textbf{bold}.}
    \label{tab:truncation-results}
    \setlength{\tabcolsep}{1.5pt}
    \renewcommand{\arraystretch}{1.25}
    \resizebox{\textwidth}{!}{%
    \begin{tabular}{l c c c c c c c}
        \toprule
        \textbf{Method} & \textbf{AIME 2024} & \textbf{AIME 2025} & \textbf{LiveCodeBench} & \textbf{GPQA} & \textbf{LiveCodeBench V4} & \textbf{OlympiadBench} & \textbf{Minerva} \\
        \midrule
        DeepSeek-R1-Distill-Qwen-7B       & 53.0          & 30.0          & 34.4          & 49.1          & 38.4          & 56.4          & 37.6          \\
        + RFT                             & 54.3          & 34.7          & 35.0          & 49.7          & 40.5          & 57.1          & 46.7          \\
        + Reflective Recovery ($t = L/2$) & 56.7          & \textbf{37.5} & \textbf{36.5} & \textbf{50.3} & \textbf{41.2} & \textbf{58.1} & \textbf{47.8}          \\
        + Reflective Recovery ($t = 3L/4$)& \textbf{60.0} & 35.8          & 35.8          & 49.0          & 38.5          & 57.1          & \textbf{47.8}          \\
        \bottomrule
    \end{tabular}
    }
\end{table*}

\paragraph{Implementation Details.}
The data curation process involves generating $N=8$ initial trajectories per problem using the base model with a temperature of $0.6$, a top-p value of $0.95$, and a maximum of 32,768 new tokens. We employ a rule-based verifier $\mathcal{V}(x, y) \in \{0, 1\}$ to check correctness against the ground truth. The verifier extracts answers from \texttt{\textbackslash boxed\{\}} expressions and compares them against ground truth using numerical tolerance ($\text{rel\_tol}=10^{-5}$) for floating-point values, supporting various mathematical expressions including fractions (\texttt{\textbackslash frac}), square roots (\texttt{\textbackslash sqrt}), and standard numerical formats. Failed trajectories are processed via our recovery re-sampling step ($K=5$) to construct $\mathcal{D}_{\text{rec}}$. For the main results, we fix the dataset size at 6k trajectories for both RFT and our method. All models are fine-tuned for 5 epochs using the Open-R1 framework~\citep{openr1} with a learning rate of $4.0 \times 10^{-5}$, a cosine schedule with warmup, and a maximum sequence length of 32,768 tokens. We use Flash Attention 2~\citep{dao2023flashattention2} for efficient training. Inference is accelerated using vLLM~\citep{kwon2023efficient}. Complete hyperparameters are provided in Appendix~\ref{app:hyperparameters}.

\paragraph{Evaluation.}
We evaluate all models on a diverse suite of challenging reasoning benchmarks, reporting zero-shot pass@1 accuracy using the lighteval framework~\citep{lighteval}. The benchmarks include AIME~\citep{ye2025aimepreview}, GPQA~\citep{rein2023gpqa}, LiveCodeBench~\citep{jain2024livecodebench}, OlympiadBench~\citep{he2024olympiadbenchchallengingbenchmarkpromoting}, and Minerva Math~\citep{lewkowycz2022solvingquantitativereasoningproblems}. 

To ensure robust evaluation, we sample 32 times per problem for AIME 2024 and AIME 2025, 16 times for LiveCodeBench, 8 times for GPQA Diamond and Minerva Math, and 1 time for OlympiadBench. This diverse suite aims to comprehensively assess the models' generalization capabilities in complex reasoning tasks.

\subsection{Main Results}
Table~\ref{tab:main-results} presents the comparative performance of our method and RFT baseline. Our Reflective Recovery approach consistently surpasses the RFT baseline across complex reasoning tasks at both the 7B and 14B scales.
For example, on the DeepSeek-R1-Distill-Qwen 7B model, our method reaches 37.5\% on AIME 2025, improving upon RFT by 2.8 points. We observe a similar improvement on Minerva, where performance increases from 46.7\% to 47.8\%.
This advantage carries over to the 14B model as well. On Minerva Math benchmark, the improvement is particularly clear, with accuracy rising from 47.4\% to 50.7\%.

Notably, despite training exclusively on mathematical reasoning problems, our method also yields a consistent gain of 2.2 points on the code benchmark LiveCodeBench, demonstrating effective cross-domain transfer. These results suggest that learning to recover from errors is a more effective training strategy than simply imitating correct solutions. By teaching the model to identify and rectify flaws in its reasoning chain, our method elicits an intrinsic self-correction capability that generalizes across different domains.

\begin{figure*}[t!]
    \centering
    \begin{subfigure}[b]{0.24\textwidth}
        \centering
        \includegraphics[width=\linewidth]{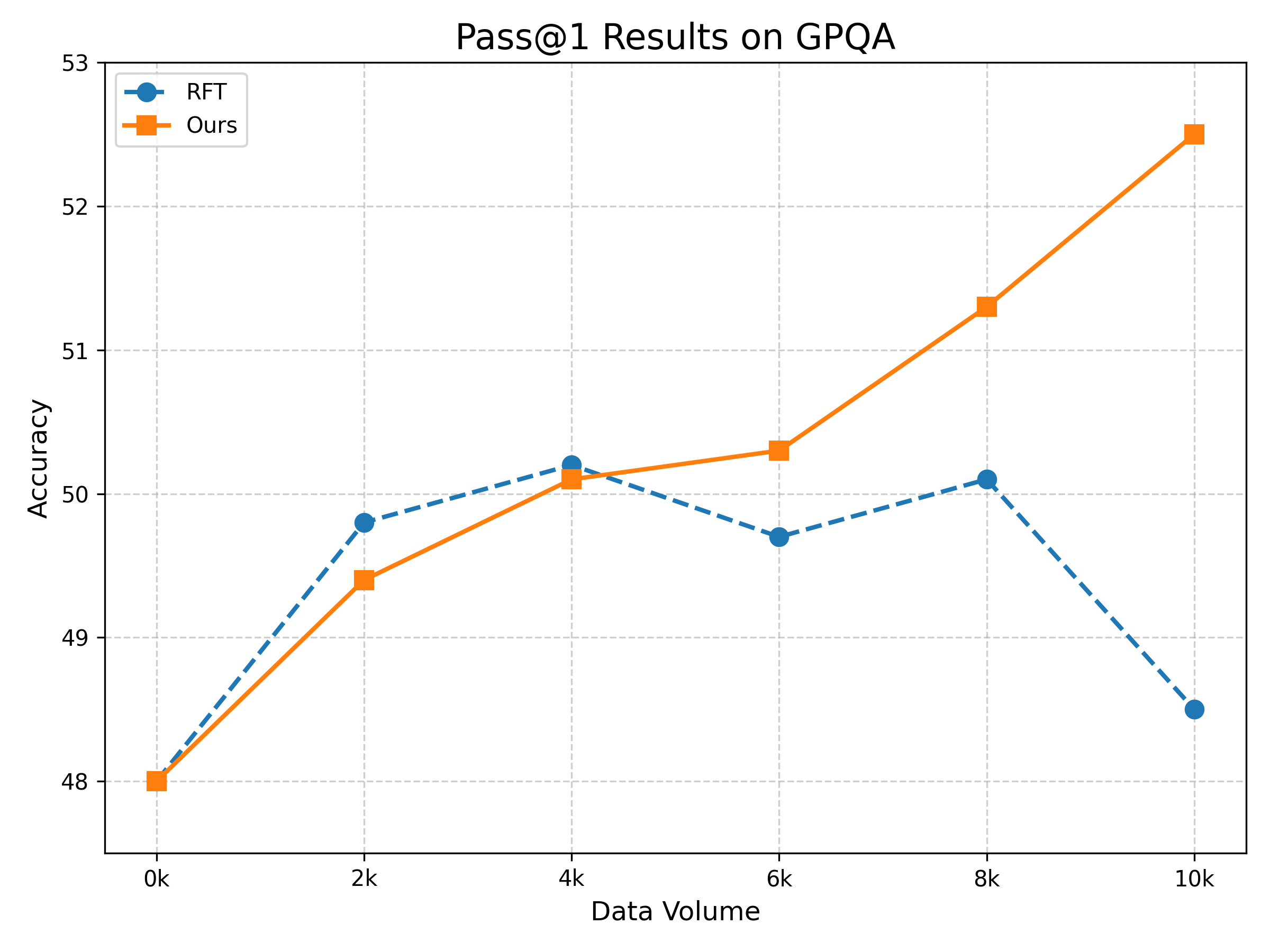}
        \caption{GPQA}
        \label{fig:dataset_a_alt}
    \end{subfigure}
    \hfill
    \begin{subfigure}[b]{0.24\textwidth}
        \centering
        \includegraphics[width=\linewidth]{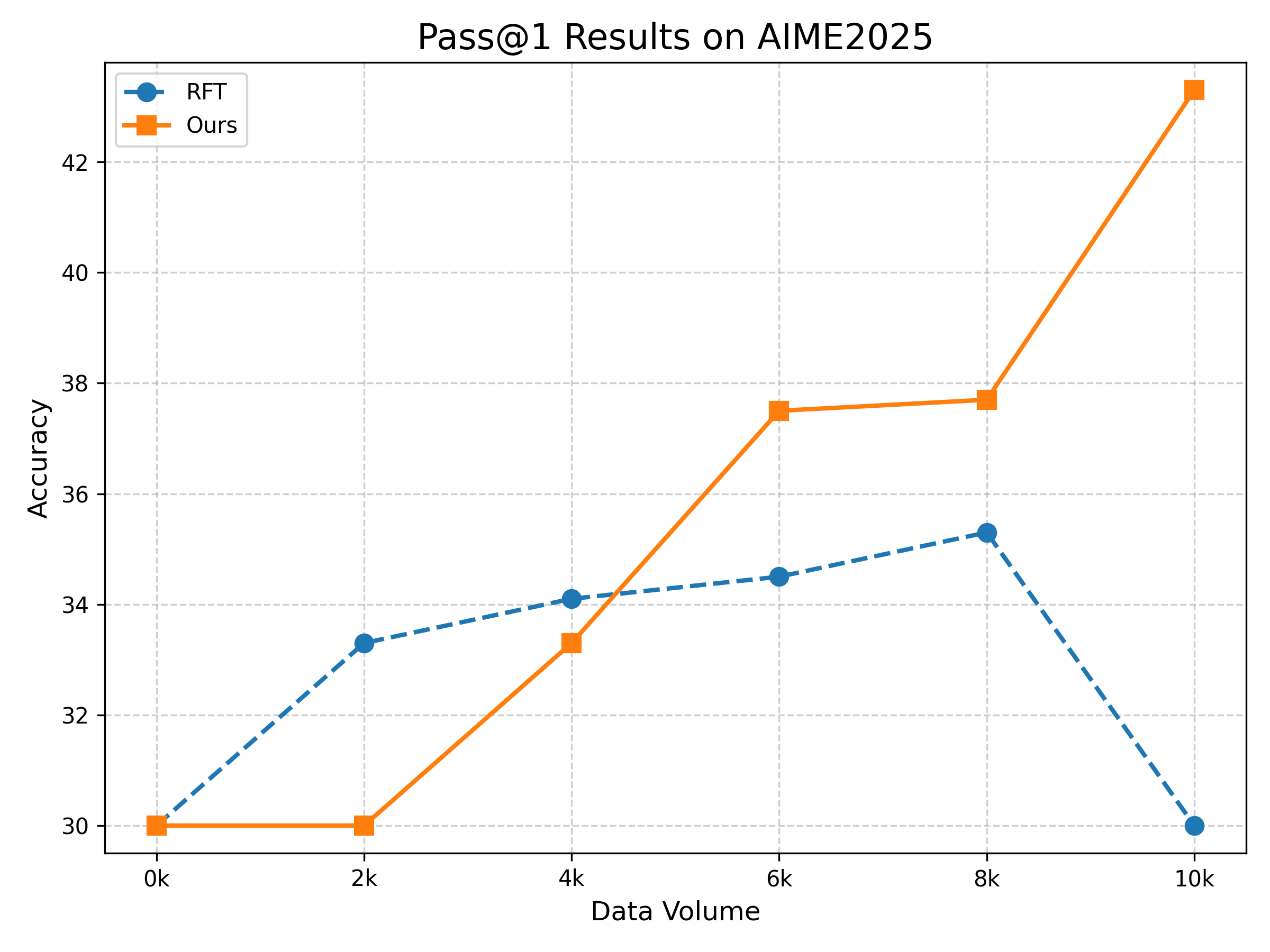}
        \caption{AIME 2025}
        \label{fig:dataset_b_alt}
    \end{subfigure}
    \hfill
    \begin{subfigure}[b]{0.24\textwidth}
        \centering
        \includegraphics[width=\linewidth]{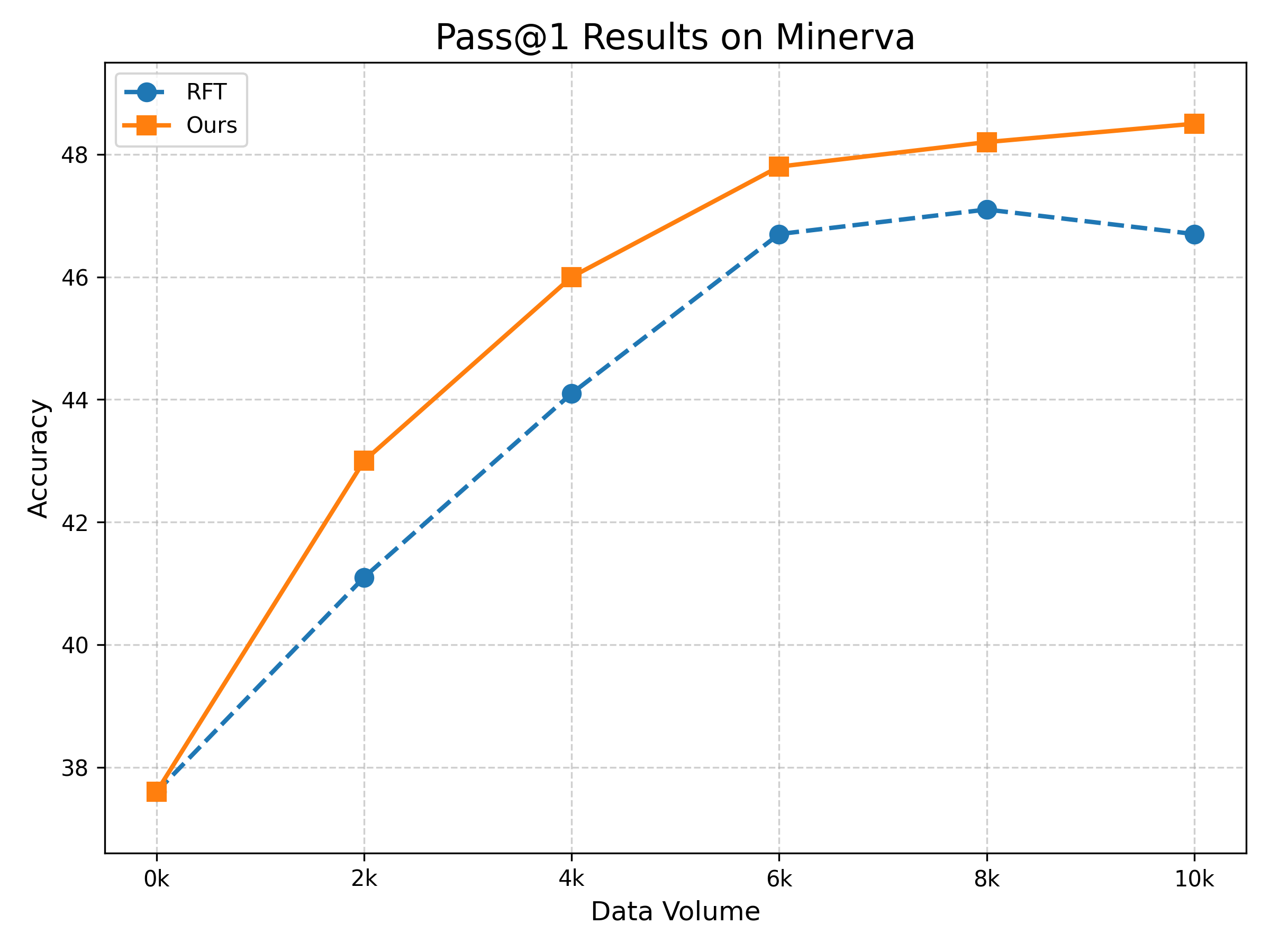}
        \caption{Minerva}
        \label{fig:dataset_d_alt}
    \end{subfigure}
    \hfill
    \begin{subfigure}[b]{0.24\textwidth}
        \centering
        \includegraphics[width=\linewidth]{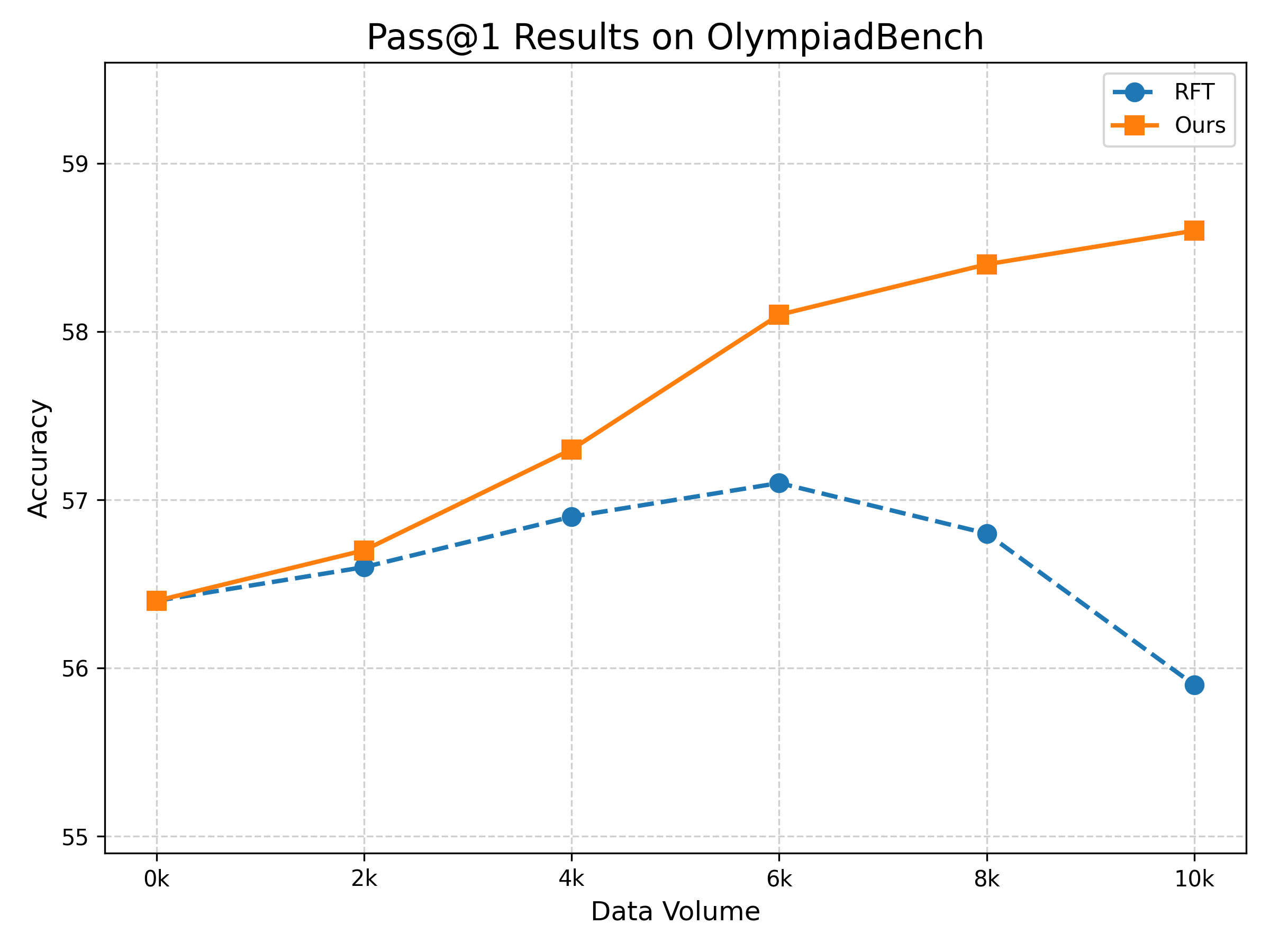}
        \caption{OlympiadBench}
        \label{fig:minera_pass1_alt}
    \end{subfigure}
    \caption{Scaling behavior of Pass@1 accuracy for our method versus RFT as data volume increases. Our method exhibits robust scaling, whereas RFT shows signs of saturation or collapse.}
    \label{fig:all_pass1_datasets_alt}
\end{figure*}

\subsection{Scaling Law Analysis}
\label{sec:scaling_analysis}

To investigate the relationship between data volume and reasoning capability, we conducted a comparative study in which both the RFT baseline and Reflective Recovery were trained on datasets ranging from 2k to 10k trajectories. The results in Figure~\ref{fig:all_pass1_datasets_alt} reveal a distinct contrast in scaling behaviors between the two approaches.

Our method shows a clear improvement and consistent scaling across most reasoning benchmarks as the amount of training data increases from 2k to 10k. For example, our method achieves a steady performance gain on Minerva, suggesting the model effectively internalizes the recovery mechanism. This indicates that our method is not just memorizing correct answers, but is learning a general ability to fix its own mistakes.


In contrast, the RFT baseline shows more fragile scaling behavior. On most benchmarks, its performance reaches a peak at intermediate data sizes, such as 4k or 8k examples, but then declines when the data volume increases to 10k. For instance, accuracy on AIME 2025 drops from 35.3\% to 30.0\% as the training data volume scales from 8k to 10k, indicating adding more data actually impairs the model's reasoning ability. Without signal provided by error recovery, RFT appears to struggle to generalize when the data distribution expands, potentially overfitting to specific solution patterns rather than learning underlying reasoning principles. Additional scaling behavior results on AIME 2024 and LiveCodeBench V4 are provided in Appendix~\ref{app:scaling}.

\subsection{Ablation Studies}

\subsubsection{Impact of Truncation Point}\label{sec:truncation_ablation}
We examine the effect of the truncation point in the failed trajectories during data curation. We compare truncating at 1/2 versus 3/4 of the Chain-of-Thought (CoT) length on DeepSeek-R1-Distill-Qwen-7B (Table~\ref{tab:truncation-results}). We observe that the 1/2 truncation generally performs better. We posit that earlier truncation (1/2) creates a more effective learning signal by forcing the model to address foundational errors in the reasoning setup. Conversely, truncating the reasoning process too late often leaves the model with a fundamentally flawed reasoning path, making it difficult to recover and introducing noise into the training signal.

\begin{table}[htbp]
\centering
\caption{Ablation study on recursive improvement. We compare models trained with a mix of Iteration 1 \& 2 data versus models trained solely on Iteration 1 data. The best results in each column are highlighted in \textbf{bold}.}
\label{tab:iteration-ablation}
\setlength{\tabcolsep}{1.5pt}
\renewcommand{\arraystretch}{1.25}
\resizebox{\columnwidth}{!}{%
\begin{tabular}{l c c c c c}
    \toprule
    \textbf{Method} & \textbf{AIME24} & \textbf{AIME25} & \textbf{LCB} & \textbf{GPQA} & \textbf{LCB\_V4} \\
    \midrule
    \multicolumn{6}{c}{\textbf{Reflective Recovery of 8k Data}} \\
    \midrule
    Iter. 1 \& 2     & \textbf{57.8} & 37.7 & \textbf{36.9} & 51.3 & \textbf{41.6} \\
    \hspace{1em}- Iter. 2        & 56.7 & \textbf{40.0} & 35.6 & \textbf{51.5} & 40.1 \\
    \midrule
    \multicolumn{6}{c}{\textbf{Reflective Recovery of 10k Data}} \\
    \midrule
    Iter. 1 \& 2     & \textbf{60.0} & \textbf{43.3} & \textbf{36.1} & \textbf{52.5} & \textbf{39.9} \\
    \hspace{1em}- Iter. 2       & 52.9 & 33.3 & 35.0 & 51.7 & 38.9 \\
    \bottomrule
\end{tabular}
}
\end{table}

\begin{figure*}[t!]
    \centering
    \includegraphics[width=0.8\textwidth]{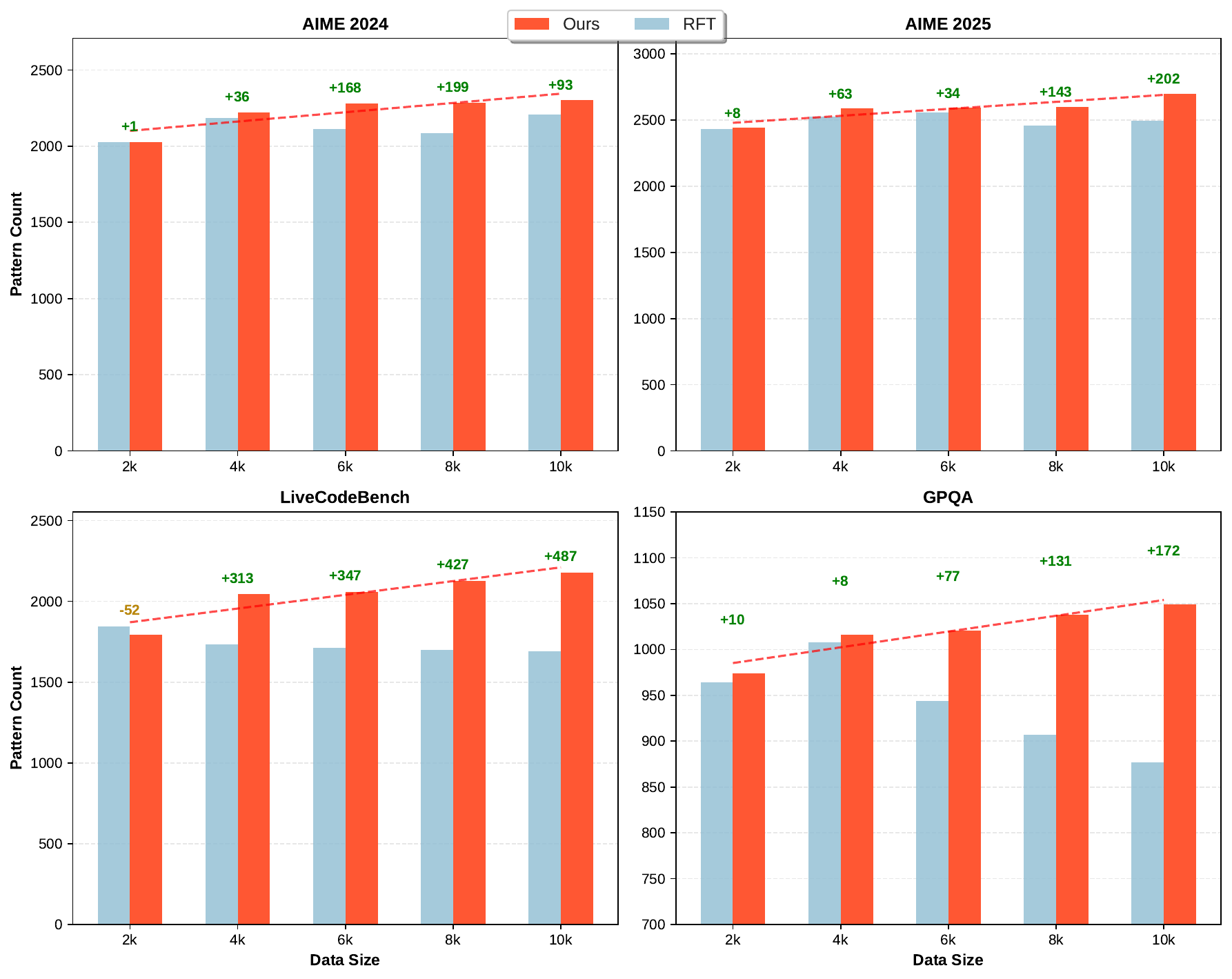}
    \vspace{-2mm}
    \caption{Frequency and distribution of "Reflective Language Patterns" generated by our model versus the RFT baseline. Our model demonstrates a notable increase.}
    \label{fig:reflective_patterns}
\end{figure*}

\subsubsection{Recursive Improvement via Iterative Curation}\label{sec:recursive_improvement}
A central question in self-evolving systems is whether the data generation process can be applied recursively to synthesize high-quality training signals. To investigate this, we generated a second iteration of data ("Iter. 2") by applying our recovery resampling process to instances that failed during the initial recovery attempt.

As shown in Table~\ref{tab:iteration-ablation}, comparing models trained with and without this second-iteration data reveals a scale-dependent effect. At the 8k data level, the inclusion of Iteration 2 data yields mixed results, implying that at lower data volumes, the model benefits more from consolidating foundational recovery skills (Iter. 1). However, at the 10k scale, the benefit of recursive data becomes clear, leading to substantial improvements across benchmarks, with gains of 7.1 points on AIME 2024 and 10.0 points on AIME 2025. These results suggest that recursive curation serves as a natural curriculum: as the model scale and data volume increase, the inclusion of more challenging recovery trajectories helps iterative self-improvement.

\subsubsection{Combining RFT and Reflective Recovery}\label{sec:synergy}

It is natural to ask whether outcome-based supervision and process-based recovery supervision can complement each other. RFT focuses on reinforcing correct reasoning trajectories from the start, while Reflective Recovery emphasizes how to recover from intermediate errors. Since they supervise different stages of reasoning, combining them may provide additional benefits.

To evaluate this possibility, we create a hybrid training set with 4k trajectories from RFT and 6k trajectories from Reflective Recovery, totaling 10k trajectories. We compare this hybrid model with the RFT-only and Reflective Recovery-only models, keeping the total data volume the same.

The results are shown in Table~\ref{tab:mix-results}. Across most datasets, the hybrid model achieves higher performance than both the RFT-only model and the Reflective Recovery-only model. For example, it achieves 62.0\% accuracy on AIME 2024 and 37.3\% on LiveCodeBench. These results indicate that combining correct reasoning demonstrations with recovery-focused supervision leads to stronger performance than either approach alone. Outcome-based supervision helps stabilize reasoning trajectories, while recovery supervision improves robustness when errors occur.

\begin{table}[htbp]
\centering
\caption{Synergy analysis. Performance of combining RFT data with our Reflective Recovery data. All models are trained on a total of 10k trajectories.}
\setlength{\tabcolsep}{1.5pt}
\renewcommand{\arraystretch}{1.25}
\label{tab:mix-results}
\resizebox{\columnwidth}{!}{%
\begin{tabular}{l c c c c c}
    \toprule
    \textbf{Method} & \textbf{AIME24} & \textbf{AIME25} & \textbf{LCB} & \textbf{GPQA} & \textbf{LCB\_V4} \\
    \midrule
    RFT                       & 52.2          & 30.0          & 34.9          & 48.5          & 39.3          \\
    Reflective Recovery       & 60.0          & 43.3          & 36.1          & \textbf{52.5} & 39.9          \\
    RFT + RR$^{\dag}$         & \textbf{62.0} & \textbf{43.5} & \textbf{37.3} & 52.0          & \textbf{41.6} \\
    \bottomrule
    \multicolumn{6}{l}{\footnotesize $^{\dag}$ RR denotes Reflective Recovery.} \\
\end{tabular}
}
\end{table}


\subsection{Analysis of Reflective Language Patterns}\label{sec:reflective_analysis}

To determine whether the performance gains arise from genuine process resilience rather than template memorization, we examined the generated reasoning traces for metacognitive markers, meaning phrases that indicate self-monitoring, doubt, or course correction. As shown in Figure~\ref{fig:reflective_patterns}, our model produces these reflective phrases more frequently and diversely than the RFT baseline.

\begin{figure}[htbp]
    \centering
    \includegraphics[width=\columnwidth]{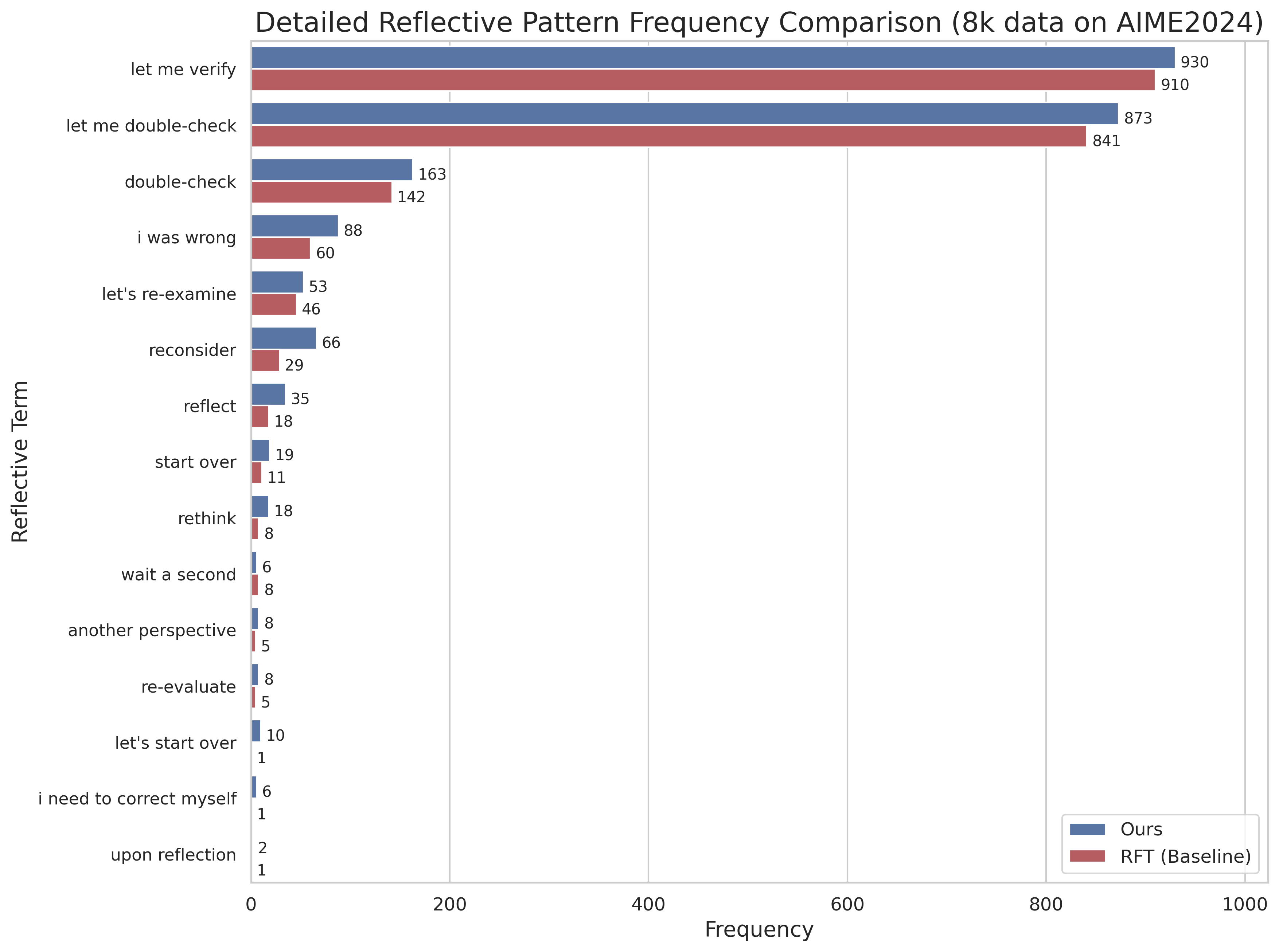}
    \caption{Detailed frequency comparison of reflective language patterns on AIME 2024 training on 8k data. The results show that our model more frequently uses phrases or language to admit mistakes and revise its reasoning trajectory, demonstrating a greater capacity for dynamic reflection and recovery.}
    \label{fig:reflective_patterns_detail}
\end{figure}

To observe this phenomenon in detail, we provide a frequency comparison on the AIME 2024 trained on 8k data in Figure~\ref{fig:reflective_patterns_detail}. Both models use common verification phrases such as "let me verify" and "let me double-check" at similarly high frequencies, but there is a clear gap in their use of substantial correction phrases. Specifically, our model explicitly stated "I was wrong" 88 times, compared to 60 in the baseline. It also used words like "reconsider" and "rethink" twice as often. For instance, "reconsider" appeared 66 times versus 29.

This behavior suggests that Reflective Recovery can actively monitor its reasoning process and recover from errors, indicating a more process-oriented reasoning capability.

\section{Conclusion}
In this paper, we introduced Reflective Recovery, a self-supervised data pipeline designed to teach language models how to fix their own mistakes during inference. Our core contribution is the demonstration that learning to recover from errors is a more reliable and scalable training strategy than simply imitating correct answers.

Our experiments on the DeepSeek-R1-Distill-Qwen 7B and 14B models show that this method outperforms outcome-base method Rejection Sampling Fine-Tuning (RFT) on challenging reasoning benchmarks. Moreover, our data scaling analysis shows that while RFT struggles or even fails with larger data volumes, our method continues to improve. Furthermore, we also observe that Reflective Recovery encourages the model to reason more actively and reflectively, exhibiting more patterns like “reconsider” or “I need to correct myself” rather than just copying answers. These findings suggest an important insight that robust reasoning is not just about avoiding errors, but about knowing how to recover from them.

\section{Limitations and Future Work}
Despite its promising results, our method faces limitations regarding computational cost and domain applicability. Creating the training data requires generating many initial failed trajectories and resampling them for recovery. Future work could explore ways to make the process more efficient, for example, by using curriculum learning to gradually introduce harder recovery tasks. We also think it would be useful to balance process-focused learning with a wider variety of data to avoid overfitting to specific domains. Finally, this recovery-oriented approach could be applied to other tasks that rely on verifiers, such as complex program synthesis or formal theorem proving.

\bibliography{anthology,custom}
\newpage
\appendix

\section{Hyperparameters and More Implementation Details}
\label{app:hyperparameters}

This section provides additional training and evaluation hyperparameters not detailed in the main text. 

\subsection{Training Configuration}
\label{app:training}

All models are trained using the Open-R1 framework. DeepSpeed ZeRO Stage 3 is used for memory-efficient distributed training.. Table~\ref{tab:training-hyperparams} summarizes the key training hyperparameters.

\begin{table}[htbp]
\centering
\caption{Training hyperparameters.}
\label{tab:training-hyperparams}
\resizebox{\columnwidth}{!}{%
\begin{tabular}{ll}
\toprule
\textbf{Hyperparameter} & \textbf{Value} \\
\midrule
Base Model & DeepSeek-R1-Distill-Qwen-7B/14B \\
Learning Rate & $4.0 \times 10^{-5}$ \\
LR Schedule & Cosine with min LR \\
Min LR Rate & 0.1 \\
Warmup Ratio & 0.03 \\
Training Epochs & 5 \\
Max Sequence Length & 32768 tokens \\
Precision & bfloat16 \\
Attention & Flash Attention \\
Max Gradient Norm & 0.2 \\
Optimizer & AdamW  \\
\bottomrule
\end{tabular}
}
\end{table}

\subsection{Evaluation Configuration}
\label{app:evaluation}

Table~\ref{tab:eval-config} provides detailed inference configuration parameters.

\begin{table}[htbp]
\centering
\caption{Inference configuration.}
\label{tab:eval-config}
\resizebox{\columnwidth}{!}{%
\begin{tabular}{ll}
\toprule
\textbf{Parameter} & \textbf{Value} \\
\midrule
Max Model Length & 32768 tokens \\
GPU Memory Utilization & 0.95 \\
Attention Backend & FlashAttention \\
Data Type & bfloat16 \\
Temperature & 0.6 \\
Top-p & 0.95 \\
\bottomrule
\end{tabular}
}
\end{table}

\section{Prompt Templates}
\label{app:prompts}

This section presents the prompt templates used for trajectory generation and model training.

\subsection{Generator Prompt}

For failure trajectory collection and recovery resampling, we use the following system prompt:

\begin{quote}
\textit{``Please reason step by step, and put your final answer within \texttt{\textbackslash boxed\{\}}."}
\end{quote}

\paragraph{Failure Trajectory Collection.} 
To generate $N$ initial trajectories, the model takes the system prompt and the problem description $x$ as input to produce a complete reasoning trajectory $y$.

\paragraph{Resampling for Recovery.} 
To generate $K$ continuation attempts, the input consists of the system prompt, the problem $x$, and the truncated trajectory $y_{<t}$, which is served as an assistant message prefix. The model then generates a continuation $\tilde{y}$ from the prefix $y_{<t}$.

During recovery resampling, prompting the model with both $x$ and $y_{<t}$ encourages it to recover from the intermediate potentially erroneous state.

\subsection{Training Data Format}

We format trajectories using a standard chat template, employing the generator prompt as the system message and the problem statement $x$ as the user message.

\begin{itemize}
    \item \textbf{RFT:} The problem $x$ serves as the sole input query. The cross-entropy loss is computed on the entire reasoning trajectory $y$.
    \item \textbf{Reflective Recovery:} The input query incorporates both the problem $x$ and a truncated prefix $y_{<t}$. Although $y_{<t}$ is formatted as the beginning part of the assistant's response, we apply loss masking to these tokens. Consequently, the model is trained exclusively on the continuation tokens $\tilde{y}$, enabling it to learn how to learn the recovery process.
\end{itemize}


\section{Additional Scaling Law Results}
\label{app:scaling}

This section presents additional scaling behavior analysis on AIME 2024 and LiveCodeBench V4 benchmarks, complementing the results shown in Figure~\ref{fig:all_pass1_datasets_alt} in Section~\ref{sec:scaling_analysis}.

\begin{figure}[t!]
    \centering
    \includegraphics[width=0.8\columnwidth]{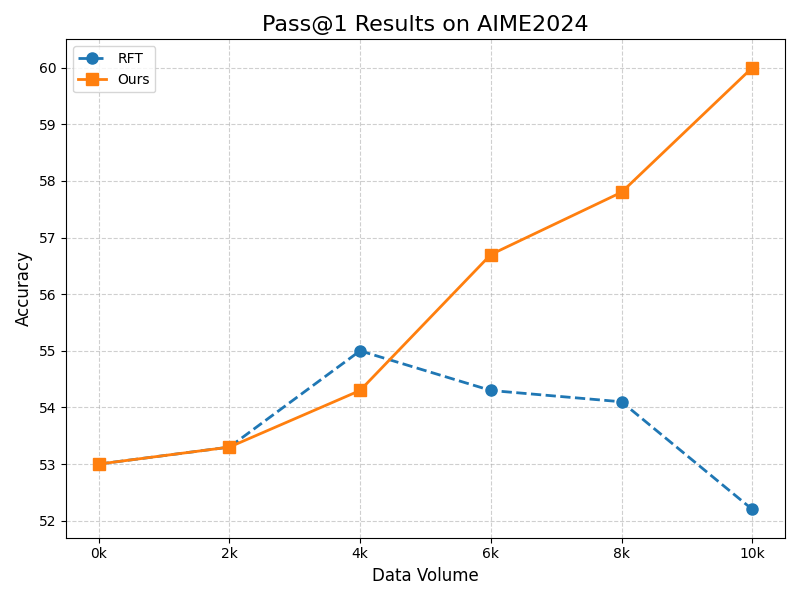}
    \caption{Scaling behavior on AIME 2024 Diamond benchmark.}
    \label{fig:scaling_gpqa}

    \vspace{1em}
    \includegraphics[width=0.8\columnwidth]{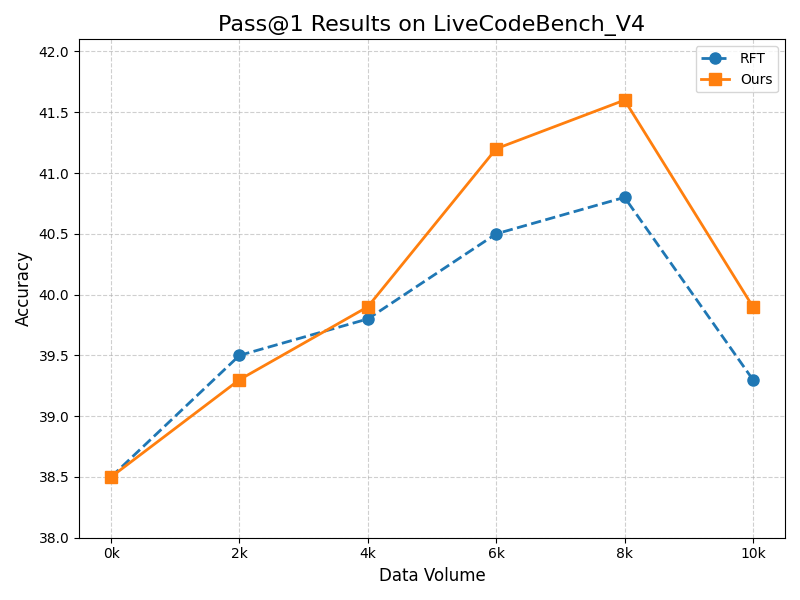}
    \caption{Scaling behavior on LiveCodeBench V4 benchmark.}
    \label{fig:scaling_lcb_v4}
\end{figure}

\subsection{Analysis of Code Benchmark Performance}

For code benchmarks, Figure~\ref{fig:scaling_lcb_v4} reveals an interesting pattern. When the data volume reaches 10k, both our method and RFT exhibit a drop in performance on LiveCodeBench V4. This may be because the model starts to overfit to the math-specific problems, focusing too much on mathematical reasoning at the cost of its general coding ability. This suggests a trade-off between domain-specific improvement and cross-domain generalization when scaling up training data from a single domain.  

\end{document}